\documentclass[10pt,twocolumn,letterpaper]{article}

\usepackage{cvpr}              %

\usepackage{graphicx}
\usepackage{amsmath}
\usepackage{amssymb}
\usepackage{booktabs}

\usepackage{graphicx, amsmath, amssymb, caption, subcaption, multirow, overpic, textpos}
\usepackage[table]{xcolor}
\usepackage[british, english, american]{babel}
\usepackage[pagebackref=false, breaklinks=true, letterpaper=true, colorlinks,
            citecolor=citecolor, linkcolor=linkcolor, bookmarks=false]{hyperref}
\definecolor{citecolor}{HTML}{0071BC}
\definecolor{linkcolor}{HTML}{ED1C24}

\usepackage{times}
\usepackage{epsfig}
\usepackage{adjustbox}
\usepackage{tabu}
\usepackage{color}
\usepackage{footnote, eucal}
\usepackage[linesnumbered, ruled]{algorithm2e}
\usepackage{algpseudocode}%
\usepackage{bm}
\usepackage{array}
\usepackage{enumitem}
\usepackage{verbatim}
\usepackage{algorithmicx}
\usepackage{comment}
\usepackage{lipsum}
\usepackage{stringstrings}
\usepackage{makecell}
\usepackage{url}
\usepackage[accsupp]{axessibility}  %

\usepackage[capitalize]{cleveref}
\crefname{section}{Sec.}{Secs.}
\Crefname{section}{Section}{Sections}
\Crefname{table}{Table}{Tables}
\crefname{table}{Tab.}{Tabs.}

\newlength\savewidth

\renewcommand{\paragraph}[1]{\vspace{1.25mm}\noindent\textbf{#1}}

\newcolumntype{x}[1]{>{\centering\arraybackslash}p{#1pt}}
\newcolumntype{y}[1]{>{\raggedright\arraybackslash}p{#1pt}}
\newcolumntype{z}[1]{>{\raggedleft\arraybackslash}p{#1pt}}

\newcommand{\app}{\raise.17ex\hbox{$\scriptstyle\sim$}}

\definecolor{deemph}{gray}{0.6}

\definecolor{baselinecolor}{gray}{.9}
\newcommand{\baseline}[1]{\cellcolor{baselinecolor}{#1}}

\def\eg{\textit{e.g.}}
\def\ie{\textit{i.e.}}

\def\vs{\textit{vs.\ }}

\newcommand{\myparagraph}[1]{{\setlength{\parskip}{0.3em} \noindent \textbf {#1}}}

\def\Our{{CAE v2}\xspace}

\makeatletter
  \newcommand\figcaption{\def\@captype{figure}\caption}
  \newcommand\tabcaption{\def\@captype{table}\caption}
\makeatother

\begin{document}

\title{\Our: Context Autoencoder with CLIP Target}

\author{Xinyu Zhang\textsuperscript{\rm 1\thanks{Equal contribution. 
}}, Jiahui Chen\textsuperscript{\rm 2,1$*$}, Junkun Yuan\textsuperscript{\rm 3,1}, Qiang Chen\textsuperscript{\rm 1}, Jian Wang\textsuperscript{\rm 1}, Xiaodi Wang\textsuperscript{\rm 1}, Shumin Han\textsuperscript{\rm 1}, \\
Xiaokang Chen\textsuperscript{\rm 4,1}, Jimin Pi\textsuperscript{\rm 1}, Kun Yao\textsuperscript{\rm 1}, Junyu Han\textsuperscript{\rm 1}, Errui Ding\textsuperscript{\rm 1}, Jingdong Wang\textsuperscript{\rm 1\thanks{
Corresponding author. 
}}\\
\textsuperscript{\rm 1}Baidu VIS~~  \textsuperscript{\rm 2}Beihang University~~ \textsuperscript{\rm 3}Zhejiang University~~ \textsuperscript{\rm 4}Peking University\\
\vspace{-0.2em}
{\tt\small \{zhangxinyu14,chenjiahui06,yuanjunkun,chenqiang13,wangjian33,wangxiaodi03,hanshumin,\}
}\\
{\tt\small \{chenxiaokang03,pijimin01,hanjunyu,dingerrui,wangjingdong\}@baidu.com
}
}

\maketitle

\begin{abstract}
Masked image modeling (MIM) learns visual representation by masking and reconstructing image patches. Applying the reconstruction supervision on the CLIP representation has been proven effective for MIM. However, it is still under-explored how CLIP supervision in MIM influences performance. To investigate strategies for refining the CLIP-targeted MIM, we study two critical elements in MIM, \ie, the supervision position and the mask ratio, and reveal two interesting perspectives, 
relying on our developed simple pipeline, context autodecoder with CLIP target (\Our).
Firstly, we observe that the supervision on visible patches achieves remarkable performance, even better than that on masked patches, where the latter is the standard format in the existing MIM methods. Secondly, the optimal mask ratio positively correlates to the model size. That is to say, the smaller the model, the lower the mask ratio needs to be. Driven by these two discoveries, our simple and concise approach \Our achieves superior performance on a series of downstream tasks. For example, a vanilla ViT-Large model achieves 81.7\% and 86.7\% top-1 accuracy on linear probing and fine-tuning on ImageNet-1K, and 55.9\% mIoU on semantic segmentation on ADE20K with the pre-training for 300 epochs. We hope our findings can be helpful guidelines for the pre-training in the MIM area, especially for the small-scale models.

\end{abstract}

\section{Introduction}
\label{sec:introduction}
Masked image modeling (MIM) \cite{bao2021beit} is at the center of self-supervised representation learning, showing good potentials on various downstream tasks, including image classification, semantic segmentation, object detection, and instance segmentation. 
It masks out some patches in the images with a pre-defined \emph{mask ratio} and adds the reconstruction supervision on a set of patches at some specific positions, \ie, \emph{supervision position}. Specifically, in most MIM methods \cite{bao2021beit, he2022masked, xie2022simmim, chen2022context}, the supervision positions are only associated with the masked patches, \ie, only adding the supervisions on the masked patches.

\par
The reconstruction loss of MIM can be applied in different domains or targets, such as 
RGB~\cite{he2022masked,xie2022simmim}, HOG~\cite{wei2022masked}, discrete visual tokens~\cite{bao2021beit,chen2022context,el2021large,peng2022beit,dong2021peco}, momentum encoders~\cite{tao2022siamese,chen2022sdae,wu2022extreme}, and pretrained models~\cite{wei2022mvp,wei2022masked}.
Recently, MVP~\cite{wei2022mvp} applies the reconstruction loss on the image representations of CLIP~\cite{radford2021learning}, \ie, minimizing the reconstruction error in the domain of CLIP representation. Benefiting from the rich multi-modality information and informative representation, the CLIP-targeted MVP performs very well.

\par
Despite that, it is still under-explored how the detailed ways of applying CLIP in MIM affect performance.
Unlike most MIM methods~\cite{bao2021beit, he2022masked, xie2022simmim, chen2022context} applying the reconstruction supervision on the masked patches, MVP supervises both masked and unmasked patches.
It raises a question: how will the \emph{supervision position} influence the CLIP-targeted MIM?
On the other hand, the \emph{mask ratio} performs differently for different supervision targets~\cite{he2022masked,bao2021beit}. 
With CLIP as the target, it is unclear how the mask ratio behaves.

\par
In this paper, we study these two critical ingredients, \ie, \textit{supervision position} and \textit{mask ratio}, in MIM with the CLIP representation as the supervision. 
To conduct the study, we develop a simple 
MIM pipeline, \ie, context autoencoder with CLIP target (\Our).
We will first analyze how the supervision position and the mask ratio influence the performance of \Our. 
Then, relying on the analyses, we implement 
\Our as a concise yet effective MIM model, producing superior performance on 
various downstream tasks. 

\par

\begin{figure*}[t!]
\centering
\includegraphics[trim =0mm 0mm 0mm 0mm, clip, width=0.85\linewidth]{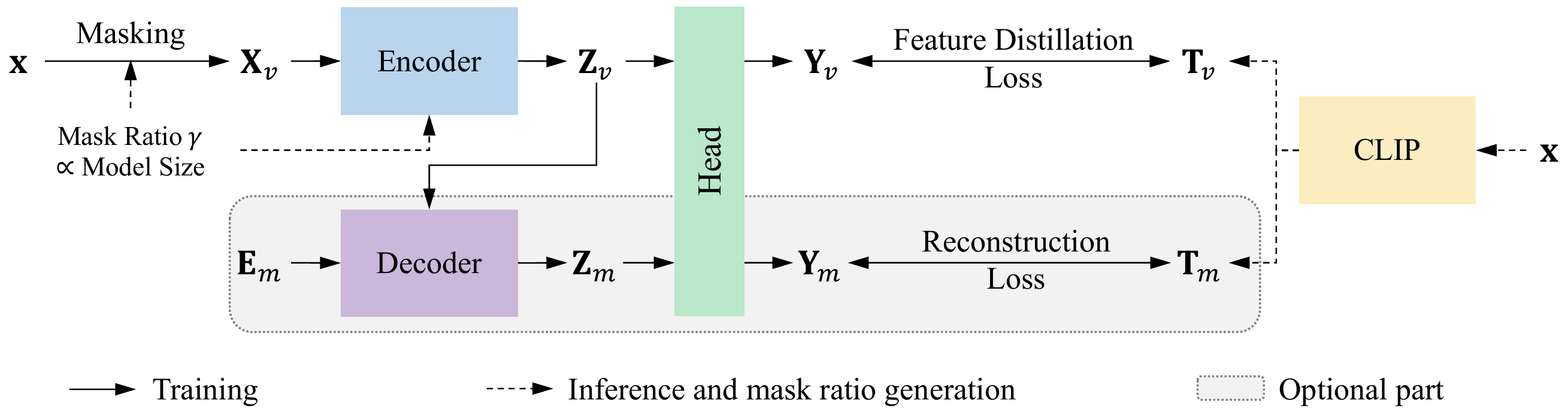}
\caption{
Overview of the proposed \Our. 
\Our first masks the input image $\pmb{\mathrm{x}}$ with the mask ratio $\gamma$, which is positively correlated with the model size of encoder.
$\propto$ represents the positive correlation.
Then, \Our inputs the visible patches $\pmb{\mathrm{X}}_v$ into the encoder to obtain the latent representation $\pmb{\mathrm{Z}}_v$.
The decoder receives $\pmb{\mathrm{Z}}_v$ and the mask token $\pmb{\mathrm{E}}_m$ to recover the latent representations of the masked patches $\pmb{\mathrm{Z}}_m$.
After a lightweight head, $\pmb{\mathrm{Z}}_v$ and $\pmb{\mathrm{Z}}_m$ are projected to $\pmb{\mathrm{Y}}_v$ and $\pmb{\mathrm{Y}}_m$.
\Our also inputs $\pmb{\mathrm{x}}$ into the CLIP model to generate the target supervisions, which are split to $\pmb{\mathrm{T}}_v$ and $\pmb{\mathrm{T}}_m$ according to the absolute positions of $\pmb{\mathrm{X}}_v$ and $\pmb{\mathrm{X}}_m$.
The optimization is applied on the prediction $\pmb{\mathrm{Y}}_v$ and the target supervision $\pmb{\mathrm{T}}_v$ of visible patches.
Meanwhile, the loss on $\pmb{\mathrm{Y}}_m$ and $\pmb{\mathrm{T}}_m$ for masked patches is optional.
}
\vspace{-1.0em}
\label{fig:framework}
\end{figure*}

First, we study on the influence of the supervision position.
Except for applying the CLIP supervision on the predictions of masked patches,
we consider to directly supervise the latent representations of visible patches with CLIP features.
Surprisingly, we find that the supervision \textit{only on visible patches} achieves remarkable performance, even better than that on masked patches.
It reveals that the visible patches can effectively extract rich semantic information from CLIP, performing like the feature distillation. 
Moreover, combining the supervisions of masked and visible patches together like \cite{wei2022mvp} brings slight performance improvement.
Therefore, we advocate that the supervision on  \textit{visible patches is a good way for the supervision position} in the CLIP-targeted MIM.

Next we explore the behavior of the mask ratio.
Recall that
MAE~\cite{he2022masked} points out a high mask ratio (75\%) is good for the balance of efficiency and effectiveness.
Others like \cite{bao2021beit,chen2022context} empirically set the mask ratio as 40-50\%, and MVP simply follows \cite{bao2021beit}.
Here we want to explore what is the optimal mask ratio in the CLIP-targeted MIM.
We conduct a series of experiments on a battery of models with different sizes and sweep the mask ratio from 15\% to 95\%.
The results are interesting, showing that the smaller models favor lower mask ratios, while larger models prefer higher ones.
This provides a new perspective that the optimal mask ratio is \textit{positively correlated with the model size}.

Based on the above analyses,
our \Our achieves superior performance on various scales of models.
Especially, with 300 epoch pre-training, \Our can boost a vanilla ViT-Large to 81.7\% and 86.7\% top-1 accuracy on linear probing and fine-tuning on ImageNet-1K, and 55.9\% mIoU on ADE20K.
We hope our analyses and findings can be useful guidelines for the future MIM studies, especially for the pre-training on lightweight models.
In summary, our contributions are:
\begin{itemize}
\itemsep -.051cm
\item We develop a simple CLIP-targeted MIM pipeline,
namely \Our, to study the supervision position and the mask ratio;
\item For the supervision position, we advocate that applying the supervision on visible patches is a good way;
\item For the mask ratio, we present that the optimal mask ratio positively correlates with the model size;
\item Driven by these analyses, 
our \Our achieves superior performance on different scales of models on various downstream tasks.
\end{itemize}

\section{\Our}
With CLIP as the supervision target, this paper aims to study the two important elements in MIM, \ie, the supervision position and the mask ratio.
We achieve this by developing a simple framework \Our, in which the supervision target is CLIP and 
the model structure is based on \cite{chen2022context,wei2022mvp} with some specific modifications.
We introduce the details as follows.

\subsection{Overview}
The overview of our \Our is illustrated in Figure~\ref{fig:framework}.
Let $\pmb{\mathrm{x}}\in\mathcal{D}$ denote an input image.
Following previous MIM methods~\cite{bao2021beit,he2022masked,wei2022masked,chen2022context}, \Our first embeds $\pmb{\mathrm{x}}$ into a total length of $N$ patches,
which are then randomly masked by a specific proportion $\gamma$.
These $N$ patches are naturally split into two non-overlapped sets, \ie, visible patches $\pmb{\mathrm{X}}_v$ and masked patches $\pmb{\mathrm{X}}_m$, where $N=\left | v \right |+\left | m \right |$.
The mask ratio is thus denoted as $\gamma=\left | m \right |/N$.
Following ~\cite{he2022masked,chen2022context}, the encoder $\mathcal{F}$ maps the visible patches $\pmb{\mathrm{X}}_v$ to the latent representations $\pmb{\mathrm{Z}}_v$.
The decoder $\mathcal{G}$ predicts the latent representations $\pmb{\mathrm{Z}}_m$ for the masked patches from mask tokens $\pmb{\mathrm{E}}_m$.
After that, the predictions of visible patches $\pmb{\mathrm{Y}}_v$ and masked patches $\pmb{\mathrm{Y}}_m$ are obtained 
via a head $\mathcal{H}$.

For the target supervision, we directly input the intact image $\pmb{\mathrm{x}}$ into the CLIP vision model $\mathcal{T}$ to generate the target supervision $\pmb{\mathrm{T}}$.
$\pmb{\mathrm{T}}$ is then split into $\pmb{\mathrm{T}}_v$ and $\pmb{\mathrm{T}}_m$ corresponding to the absolute positions of $\pmb{\mathrm{X}}_v$ and $\pmb{\mathrm{X}}_m$.
The optimization is applied on $\pmb{\mathrm{Y}}_v$ and $\pmb{\mathrm{T}}_v$, 
and we also study to add the supervision on $\pmb{\mathrm{Y}}_m$ and $\pmb{\mathrm{T}}_m$.

\subsection{Architecture}
\Our contains four modules, \ie, one encoder, one decoder, one MIM head, and one CLIP model.

\myparagraph{Encoder.}
The encoder $\mathcal{F}$ only receives the visible patches $\pmb{\mathrm{X}}_v$ following \cite{he2022masked,chen2022context}.
$\mathcal{F}$ maps the visible patches $\pmb{\mathrm{X}}_v$ to the latent representations $\pmb{\mathrm{Z}}_v$ across a stack of transformer blocks.
The operation of $\mathcal{F}$ is based on self-attention.
In this paper, we utilize a series of ViTs~\cite{dosovitskiy2020image} to form the encoder, including ViT-Tiny, ViT-Small, ViT-Base, and ViT-Large.

\myparagraph{Decoder.} The decoder $\mathcal{G}$ predicts the latent representation $\pmb{\mathrm{Z}}_m$ for the masked patches from the mask tokens $\pmb{\mathrm{E}}_m$, conditioned on the
visible latent representation $\pmb{\mathrm{Z}}_v$.
It is inspired by the latent contextual regressor in CAE~\cite{chen2022context}.
$\mathcal{G}$ performs as the same as cross-attention.
Here, we utilizes a lighweight decoder in \Our, \ie, one-layer transformer block\footnote{We find that one-layer transformer block as the decoder ready can produce promising performance. For parameter-friendly, we use this structure as the default setting.
We do not analyze the influence of the depth of decoder, since it is not the main concern in this paper.
}.

\myparagraph{Head.} The head $\mathcal{H}$ maps the latent predictions $\pmb{\mathrm{Z}}_v$ and the latent representations $\pmb{\mathrm{Z}}_m$ to $\pmb{\mathrm{Y}}_v$ and $\pmb{\mathrm{Y}}_m$, respectively.
$\pmb{\mathrm{Y}}_v$ and $\pmb{\mathrm{Y}}_m$ share the same form with the target supervisions $\pmb{\mathrm{T}}_v$ and $\pmb{\mathrm{T}}_m$.
In this work, we only use a FC (fully-connected) layer followed by a LN (layernorm) layer in $\mathcal{H}$.

\myparagraph{CLIP model.} The vision branch of CLIP model $\mathcal{T}$ generates the target supervisions $\pmb{\mathrm{T}}$.
In the whole paper, we only use the ViT-Base of the CLIP model as $\mathcal{T}$.
$\pmb{\mathrm{T}}$ is then split into the target supervisions for visible patches $\pmb{\mathrm{T}}_v$ and for masked patches $\pmb{\mathrm{T}}_m$ according to the absolute positions of $\pmb{\mathrm{X}}_v$ and $\pmb{\mathrm{X}}_m$.

\subsection{Study, Discovery and Analysis}
\label{sec:study}
We pay attention to two critical elements in MIM, \ie, the supervision position and the mask ratio.

\myparagraph{Supervision position.}
Most previous MIM methods~\cite{he2022masked,chen2022context,bao2021beit} apply the reconstruction supervision on the predictions of masked patches.
With CLIP as the target, MVP~\cite{wei2022mvp} supervises both visible and masked patches.
Here, we do experiments to study how will the
supervision position influence the CLIP-targeted MIM.

We systematically analyze three kinds of supervision positions based on \Our, \ie, only on the predictions of visible patches $\pmb{\mathrm{Y}}_m$, only on the predictions of masked patches $\pmb{\mathrm{Y}}_v$, and on both $\pmb{\mathrm{Y}}_v$ and $\pmb{\mathrm{Y}}_m$.
The loss function can be formulated as follows:
\begin{equation}
\centering
\begin{aligned}
L=\frac{\delta_{v}\cdot \ell(\pmb{\mathrm{Y}}_{v},\pmb{\mathrm{T}}_{v}) + \delta_{m}\cdot \ell(\pmb{\mathrm{Y}}_{m},\pmb{\mathrm{T}}_{m})}
{\delta_{v}\cdot \left | v \right | + \delta_{m}\cdot \left | m \right |},
\end{aligned}
\label{eq:loss}
\end{equation} 
where $\ell$ is the loss function.
By default, we use the cosine distance as the loss function.
$\delta_{v}$/$\delta_{m}$ is the indicative function, controlling whether to use visible/masked patches for optimization.
If $\delta_{m}=1-\delta_{v}=1$, Eq.~\eqref{eq:loss} reduces to only use the reconstruction loss for masked patches as in \cite{he2022masked,chen2022context,bao2021beit}.
If $\lambda_{m}=1-\delta_{v}=0$, the loss becomes the feature distillation loss for visible patches.
When setting $\lambda_{m}=\delta_{v}=1$, the optimization is imposed on both visible and masked patches as in \cite{wei2022mvp}.

From the experiments (see Table~\ref{tab:supervision_position}), we observe an interesting phenomenon: \textit{only applying the supervision on visible patches can achieve remarkable performance}.
We conjecture this benefit mainly comes from the powerful CLIP model by knowledge distillation.
Intriguingly, the distillation loss works well on a subset of image patches, in which some contextual information are missing.
Moreover, despite the relatively inferior performance when only using the reconstruction supervision on masked patches,
it bring slightly positive improvement when
combining the reconstruction supervision on the masked patches with the feature distillation loss on the visible patches.
We believe that in this way, the supervision on masked patches performs as a regularization for the representation learning, which is beneficial to alleviating the over-fitting when training  only on visible patches.

Our findings are different from the common sense in the current MIM methods~\cite{bao2021beit,he2022masked,chen2022context} that only compute the loss on the masked patches, which
is inherited from BERT~\cite{bert} in the NLP areal and has been verified by most current works.
Therefore, in CLIP-targeted MIM, we provide a new perspective that the feature distillation on the partial image patches (here referred to visible patches) is a good choice for the model optimization.

\begin{figure}[t]
\centering
\includegraphics[trim =0mm 0mm 0mm 0mm, clip, width=1.0\linewidth]{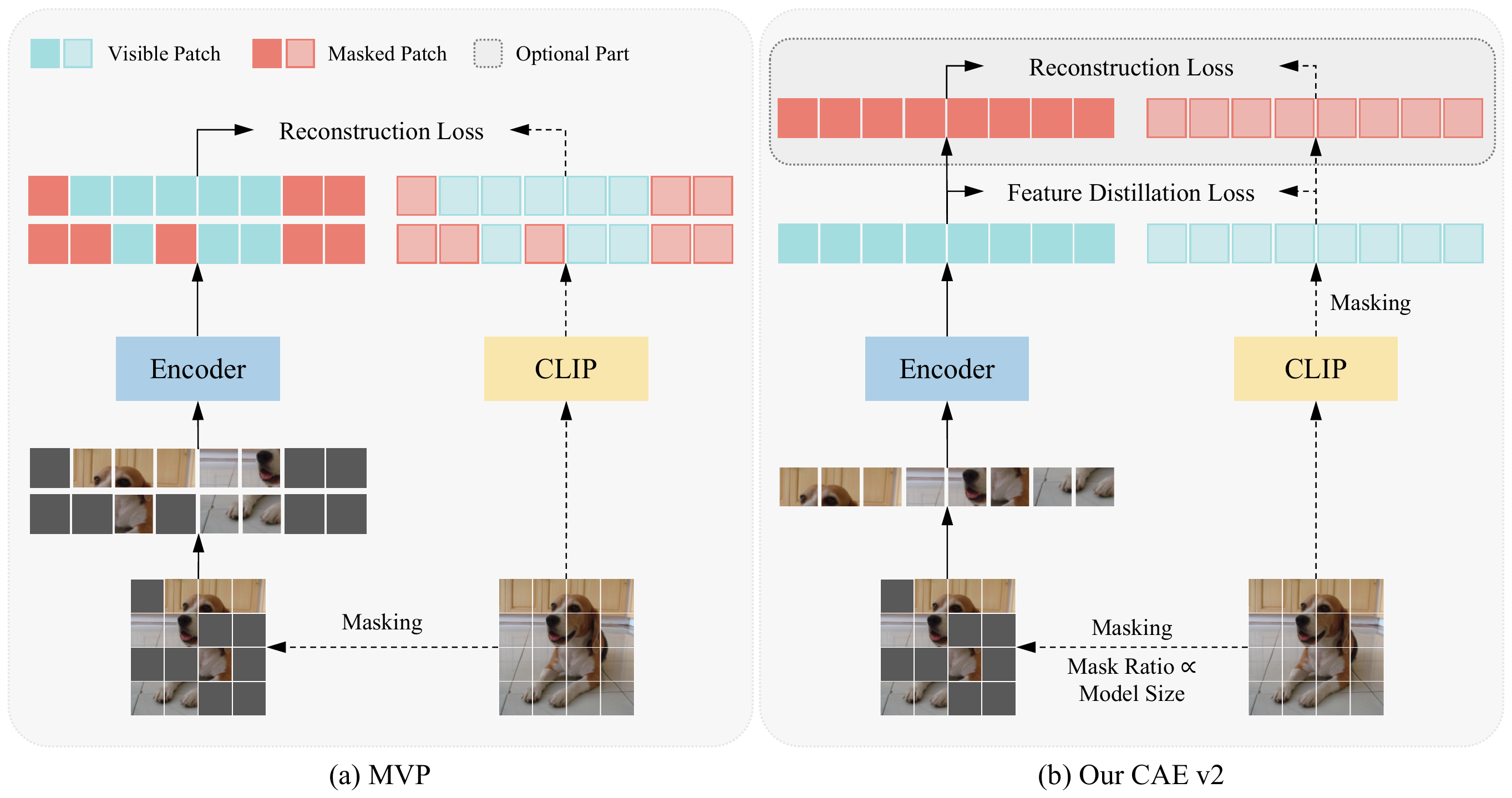}
\caption{
MVP~\cite{wei2022mvp} \vs our \Our. We mainly study the supervision position and the mask ratio in the CLIP-targeted MIM.
}
\vspace{-0.5cm}
\label{fig:comparison_mvp}
\end{figure}

\myparagraph{Mask ratio.}
The MIM methods generally mask a specific percentage of patches on $\pmb{\mathrm{x}}$ as the input for the model training.
For example, \cite{he2022masked} utilizes the mask ratio of 75\%, while
~\cite{bao2021beit,wei2022mvp} and \cite{chen2022context} empirically set the mask ratio as 40\% and 50\%.
Driven by the above experience that the CLIP supervision on visible patches can achieve good results in our \Our, we naturally consider to study what is the optimal option for the mask ratio.

We begin from a high mask ratio $\gamma$, \ie, 75\% as in \cite{he2022masked}.
We find that despite the reasonable performance on the large model (\eg, ViT-Base), it performs less than satisfactory on smaller models like ViT-Tiny/Small (see Figure~\ref{fig:mask_ratio}).
We thus gradually decrease the mask ratio, leading to more visible patches.
The performance on all scales of models starts to improve at the beginning of reducing the mask ratio.
This trend is especially true for the downstream task like semantic segmentation, 
verifying that the high mask ratio is not necessary in the CLIP-targeted MIM.

With the continued decreas of the mask ratio, the models with different sizes perform differently.
We observe that the smaller models favor lower proportions of mask ratios, while the larger models prefer relatively higher ones.
The underlying reason may be that it is hard for the small-size model to optimize on a small subset of patches where most contextual information are missing.
So it is better to reduce the difficulty by using more visible patches, \ie, a lower mask ratio.
On the contrary, large-scale models learn representations from plenty of patches easily.
Therefore, a high mask ratio can make MIM harder to ease the over-fitting. 

Based on the above observation, we point out that \textit{the optimal mask ratio is positively correlated with the model size}.
That is to say, the larger the model, the higher the mask ratio, and preferring more challenging work; otherwise, the smaller model favors a lower mask  ratio.
We believe that this discovery can be a useful guideline in the MIM pre-training area, especially for the small models.

\begin{table}
\small
\begin{center}
\setlength{\tabcolsep}{3.0mm}
\renewcommand{\arraystretch}{1.1}
\scalebox{1.0}{
\begin{tabular}{l|cc|c|c|c}
\hline
\multirow{2}{*}{Model} & \multicolumn{2}{c|}{Supervision} & \multicolumn{2}{c|}{IN-1K} & \multicolumn{1}{c}{ADE20K} \\
\cline{2-6}
 & \multicolumn{1}{c}{$\pmb{\mathrm{Y}}_m$} & \multicolumn{1}{c|}{$\pmb{\mathrm{Y}}_v$} & LIN & FT & mIoU \\ 
\hline
\hline
\multirow{3}{*}{ViT-Tiny} & \checkmark & - & 64.9 & 77.2 & 44.1 \\
& - & \checkmark & 68.8 & 77.4 & 44.2 \\
  & \baseline{\checkmark} & \baseline{\checkmark} & \baseline{\textbf{69.3}} & \baseline{\textbf{77.8}} & \baseline{\textbf{44.7}}  \\
\hline
\multirow{3}{*}{ViT-Small} & \checkmark & - & 73.9 & 82.4 & 49.6  \\
& - & \checkmark & 77.3 & \textbf{82.8} & 49.1 \\
  & \baseline{\checkmark} & \baseline{\checkmark} & \baseline{\textbf{77.5}} & \baseline{82.7} & \baseline{\textbf{49.7}}  \\
\hline
\multirow{3}{*}{ViT-Base} &  \checkmark & - & 78.4 & 85.0 & 52.7 \\
& - & \checkmark & 80.5 & 85.2 & \textbf{53.1} \\
  & \baseline{\checkmark} & \baseline{\checkmark} & \baseline{\textbf{80.6}} & \baseline{\textbf{85.3}} & \baseline{52.9} \\
\hline
\end{tabular}
}
\end{center}
\vspace{-0.5cm}
\caption{Influences of the supervision position in our \Our. Default settings are marked in \colorbox{baselinecolor}{gray}.}
\vspace{-0.5cm}
\label{tab:supervision_position}
\end{table}

\myparagraph{Discussion.}
The most relevant work for ours is MVP~\cite{wei2022mvp}.
It is noted that our \Our is not contradictory to MVP, even though we both use CLIP as the target. 
We focus on the analyses under the CLIP-targeted MIM and propose new perspectives, while MVP highlights the rich information brought by the CLIP target. 
Specifically,
in this work, we go further one step to study two important ingredients in MIM, including the supervision target and the mask ratio.
First, we find only applying the CLIP supervision on visible patches already achieves comparable or even superior performance compared to the optimization on all patches as in MVP.
Second, MVP inherits the mask ratio (40\%) from \cite{bao2021beit} and applies it on both base- and large-size models.
Differently, we study the effect of the mask ratios on a battery of models with different scales, and find the mask ratio is highly related to the model size.
In addition, our \Our with these two findings demonstrates large performance gains. 
We illustrate the detailed paradigms of our \Our and MVP in Figure~\ref{fig:comparison_mvp} for better comparisons.

\section{Experiments}
\label{sec:experiments}

\subsection{Settings}
\myparagraph{Model structures.} We study a series of vision transformer backbones~\cite{dosovitskiy2020image}, including ViT-Tiny (12 layers with $dim$$=$192), ViT-Small (12 layers with $dim$$=$384), ViT-Base (12 layers with $dim$$=$768), and ViT-Large (24 layers with $dim$$=$1024).
Note that for ViT-Tiny, we follow \cite{wang2022closer} to increase the number of heads from 3 to 12, which gives better results on ImageNet-1K~\cite{deng2009imagenet}. For other models, we strictly follow the model configurations as in~\cite{dosovitskiy2020image}.

To eliminate the influence of different sizes of CLIP models, we adopt the vision branch ViT-Base/16 of CLIP\footnote{The official pre-trained CLIP model is available at \url{https://github.com/openai/CLIP/blob/main/clip/clip.py}.} as the target for all pre-training experiments with ViT-Tiny/Small/Base/Large~\cite{dosovitskiy2020image}.

\begin{table}
\small
\begin{center}
\renewcommand{\arraystretch}{1.1}
\setlength{\tabcolsep}{2.7mm}{
\begin{tabular}{l|c|c|c|c}
\hline
\multirow{2}{*}{Model} & \multirow{2}{*}{Loss type} & \multicolumn{2}{c|}{IN-1K} & \multicolumn{1}{c}{ADE20K} \\
\cline{3-5}
 &  & LIN & FT & mIoU \\ 
\hline
\hline
\multirow{3}{*}{ViT-Tiny} & MSE & 69.1 & 77.3 & \textbf{44.8} \\
 & Smooth-$l$1 & \textbf{69.4} & 77.6 & 43.7 \\
  & \baseline{Cosine distance} & \baseline{69.3} & \baseline{\textbf{77.8}} & \baseline{44.7} \\
\hline
\multirow{3}{*}{ViT-Small} & MSE & 77.3 & 82.7 & \textbf{49.8} \\
 & Smooth-$l$1 & 77.4 & \textbf{82.8} & \textbf{49.8} \\
  & \baseline{Cosine distance} & \baseline{\textbf{77.5}} & \baseline{82.7} & \baseline{49.7} \\
\hline
\multirow{3}{*}{ViT-Base} & MSE & 80.4 & \textbf{85.3} & \textbf{52.9} \\
 & Smooth-$l$1 & 80.5 & 85.2 & 52.0\\
  & \baseline{Cosine distance} & \baseline{\textbf{80.6}} & \baseline{\textbf{85.3}} & \baseline{\textbf{52.9}} \\
\hline
\end{tabular}}
\end{center}
\vspace{-0.5cm}
\caption{Ablation on the loss type in our \Our. 
We use the cosine distance as the default loss function (marked in \colorbox{baselinecolor}{gray}).
}
\vspace{-0.3cm}
\label{tab:loss_type}
\end{table}

\myparagraph{Pre-training.}
Following most previous MIM methods~\cite{bao2021beit,he2022masked,chen2022context,wei2022mvp,wang2022closer}, 
we use ImageNet-1K (IN-1K) dataset~\cite{deng2009imagenet} for all pre-training experiments.
The input images are with the size of $224\times 224$ and partitioned into $14\times 14$ patches with the patch size being $16\times 16$.
We apply random resized cropping and horizontal flipping during pre-training.

The pre-training settings are almost the same as CAE~\cite{chen2022context}, except for the mask ratios that are analyzed in Section.~\ref{sec:ablation}.
By default, we use $15\%$, $25\%$, $50\%$, and $50\%$ mask ratios on ViT-Tiny, ViT-Small, ViT-Base, and ViT-Large, respectively.
We use AdamW~\cite{Loshchilov2019} for optimization and train the \Our for 300 epochs for all scales of ViTs~\cite{dosovitskiy2020image}.
The detailed pre-training settings are shown in the supplementary material.

\begin{figure*}[t]\centering
\includegraphics[width=.95\linewidth]{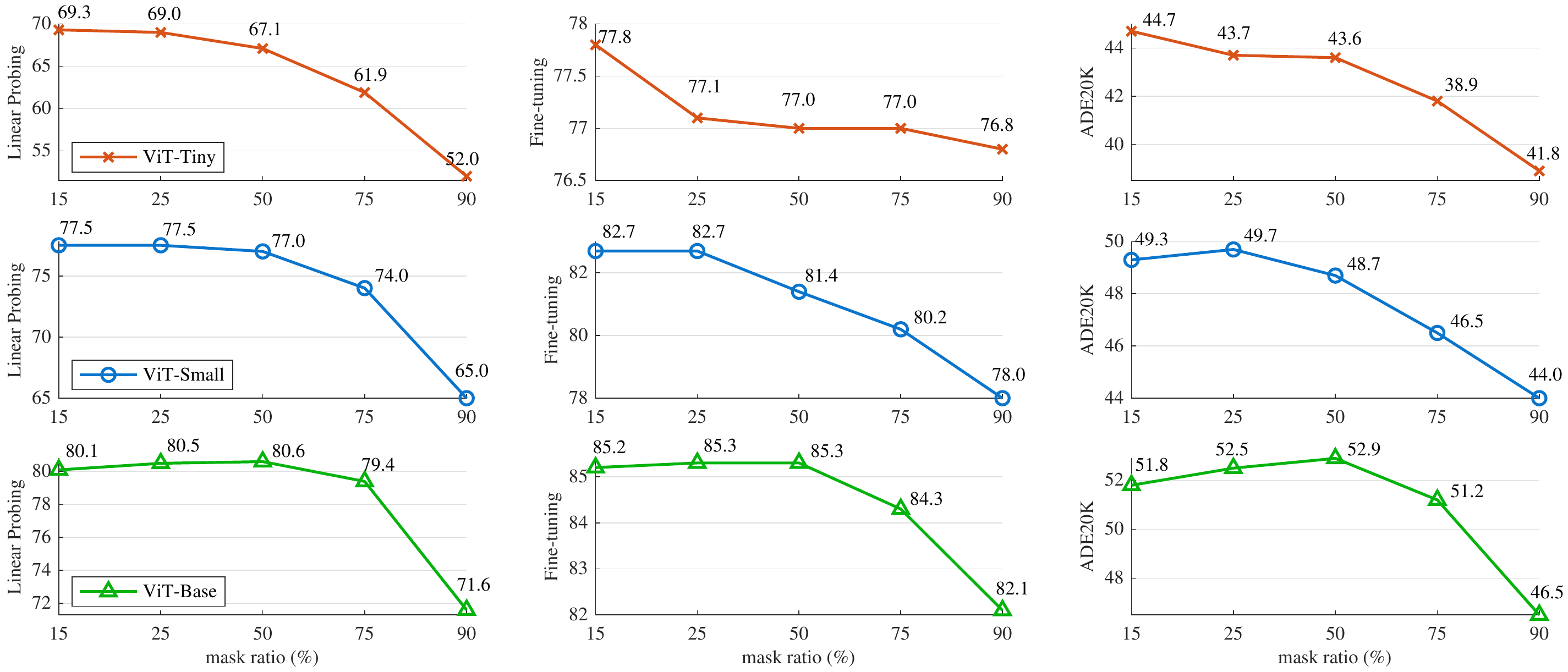}
\vspace{-.7em}
\caption{Influences of the mask ratio in our \Our on different model sizes, including (top row) ViT-Tiny, (middle row) ViT-Small and (bottom row) ViT-Base. The optimal mask ratio is positively correlated to the model size.
A higher mask ratio is more appropriate to a larger model, while the smaller model prefers a lower mask ratio.
The y-axes is the Top-1 accuracy (\%) on (left column) linear probing and (middle column) fine-tuning on ImageNet-1K, and (right column) mIoU (\%) on ADE20K.
}
\label{fig:mask_ratio}
\vspace{-1em}
\end{figure*}

\myparagraph{Evaluation.}
We evaluate our \Our on various downstream tasks.
For image classification, we conduct evaluations on ImageNet-1K~\cite{deng2009imagenet} with both linear probing (LIN) and fine-tuning (FT) protocols.
Without specification, in all experiments, we conduct the linear probing for 90 epochs and the fine-tuning for 100 epochs.
For semantic segmentation, we follow BEiT\cite{bao2021beit} to use UperNet~\cite{xiao2018unified} and report the mIoU on ADE20K~\cite{zhou2017scene} dataset.
For objection detection and instance segmentation, we use COCO~\cite{lin2014microsoft} as the evaluation dataset.
We adopt both Mask R-CNN~\cite{he2017mask} and Cascade Mask R-CNN~\cite{cai2018cascade} frameworks and report $\text{AP}^{b}$ and $\text{AP}^{m}$ on the COCO val split.
Please refer to the supplementary material for more training details on various downstream tasks.

\subsection{Main Properties}
\label{sec:ablation}
We mainly explore two critical ingredients, the supervision position and the mask ratio, in \Our when using CLIP as the supervision target. Compared with previous pre-training methods, we observe different properties and trends. In addition, we also give investigations on the loss types and the masking types. Details are given below.

\myparagraph{Supervision position.} Modern pre-training methods~\cite{bao2021beit,he2022masked,chen2022context} only give supervision on masked patches, as they find that learning with visible patches is an easy task and may leak information, result in trivial solutions, and degenerate the representation learning. When using CLIP~\cite{radford2021learning} as the pre-training target, the situation is different. 

Table~\ref{tab:supervision_position} provides the detailed results of the influence of the supervision position on ImageNet-1K~\cite{deng2009imagenet} and ADE20K~\cite{zhou2017scene} based on \Our. 
Different from existing MIM methods~\cite{bao2021beit,he2022masked,chen2022context}, we observe that only adding supervision on visible patches already presents remarkable results on both linear probing, image classification fine-tuning, and semantic segmentation fine-tuning. While only adding supervision on masked patches~\cite{bao2021beit,he2022masked,chen2022context} shows worse performance with linear probing (64.9\% \textit{vs.} 68.8\% with ViT-Tiny, 73.9\% \textit{vs.} 77.3\% with ViT-Small, and 78.4\% \textit{vs.} 80.5\% with ViT-Base), but competitive results with image classification fine-tuning and semantic segmentation fine-tuning (Table~\ref{tab:supervision_position}). We conjecture that the rich semantics in CLIP~\cite{radford2021learning} make the supervision on visible patches
work
as a knowledge distillation task. It is different from previous methods where the semantics in the target is not precise enough to provide strong supervision on visible patches. Moreover, by combining the supervision on visible patches and masked patches (the rows marked with \colorbox{baselinecolor}{gray} in Table~\ref{tab:supervision_position}), we obtain a slight improvement over the one only with supervision on visible patches. Based on the observation, 
the supervision on masked patches
in our \Our can be considered as a regularization for representation learning with CLIP as the target.

\begin{figure}[t]
\centering
\includegraphics[trim =0mm 0mm 0mm 0mm, clip, width=1.0\linewidth]{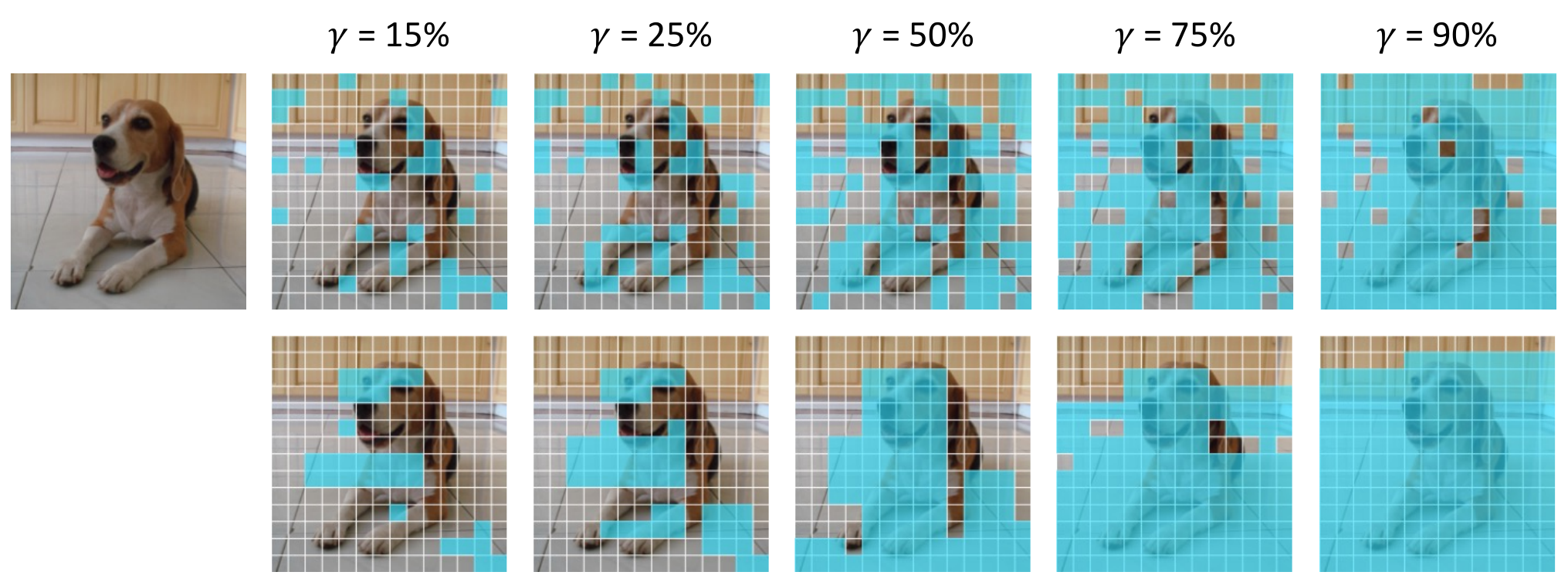} \\
\setlength{\belowcaptionskip}{0.3cm}  %
\figcaption{
Illustration of corrupted images with different mask ratios $\gamma$ via (top row) block-wise sampling strategy (our default) and (bottom row) random sampling strategy.
}
\vspace{-0.8cm}
\label{fig:sampling_strategy}
\end{figure}

\myparagraph{Mask ratio.}
Given that 
the supervision on masked patches is a regularization for the learning of \Our, we suppose that it may not be appropriate to adopt a high mask ratio (75\% in MAE~\cite{he2022masked}, 40\%-50\% in BEiT~\cite{bao2021beit}, CAE~\cite{chen2022context}, and MVP~\cite{wei2022mvp}) for all scales of ViTs. 
We study different mask ratios, ranging from 15\% to 90\% (see Figure~\ref{fig:sampling_strategy}), for a series of ViTs (Tiny, Small, and Base) with our \Our.

Figure~\ref{fig:mask_ratio} gives the performance of ViT-Tiny/ViT-Small/ViT-Base~\cite{dosovitskiy2020image} under various mask ratios using three evaluation methods: linear probing, image classification fine-tuning, and semantic segmentation fine-tuning. The results give a clear trend on the choice of mask ratios, \ie, {\em the best mask ratio is positively related to the model size}. From the curves in Figure~\ref{fig:mask_ratio}, we also find that with the selected mask ratio exceeding the best mask ratio, the performances on downstream tasks drop quickly, especially fine-tuning. The evidence indicates the significance of our study on the mask ratio and provides a guideline for choosing mask ratios for different scales of ViTs.
According to the experiments in Figure~\ref{fig:mask_ratio}, we use a mask ratio of $15\%$ for ViT-Tiny, $25\%$ for ViT-Small, and $50\%$ for ViT-Base.

\myparagraph{Loss type.} As the target of \Our is the CLIP features, we explore the influence of different loss functions in the feature space.
Table~\ref{tab:loss_type} shows the results of \Our on different scales of models with different loss functions, including the mean square error (MSE), Smooth-$l$1 and the cosine distance. We only observe slight differences ($\leq$$0.5\%$) in performance on ImageNet1K~\cite{deng2009imagenet}. On ADE20K~\cite{zhou2017scene}, the variances are slightly bigger, indicating the loss function of using CLIP as the target gives more impact on challenging downstream tasks. We adopt the cosine distance loss as the default choice.

\myparagraph{Mask sampling strategy.} We also compare different mask sampling strategies in our \Our, \ie, random sampling~\cite{he2022masked} and block-wise sampling~\cite{bao2021beit,chen2022context,wei2022mvp} (as shown in Figure~\ref{fig:sampling_strategy}). There are only $\sim$$0.1\%$ gaps between these two sampling strategies on linear probing and image classification fine-tuning (see Table~\ref{tab:mask_strategy}). When it comes to semantic segmentation, the block-wise sampling shows better performance than the random sampling. In \Our, we use block-wise sampling strategy by default.

\subsection{Main Results}
Based on the analyses of the supervision position in Section~\ref{sec:study}, we find that only applying the feature distillation supervision on visible patches can achieve good performance.
We denote this format as \Our.
Meanwhile, adding the reconstruction supervision on masked patches can further improve the performance, which is denoted as \Our$+$.
We report both \Our and \Our$+$ in this subsection.

\begin{table}[t]
\small
\begin{center}
\renewcommand{\arraystretch}{1.15}
\setlength{\tabcolsep}{2.5mm}{
\begin{tabular}{l|c|c|c|c}
\hline
\multirow{2}{*}{Model} & \multirow{2}{*}{Mask strategy} & \multicolumn{2}{c|}{IN-1K} & \multicolumn{1}{c}{ADE20K} \\
\cline{3-5}
 &  & LIN & FT & mIoU \\ 
\hline
\hline
\multirow{2}{*}{ViT-Tiny} & Random & 69.1 & 77.4 & 43.8 \\
  & \baseline{Blockwise} & \baseline{\textbf{69.3}} & \baseline{\textbf{77.8}} & \baseline{\textbf{44.7}} \\
\hline
\multirow{2}{*}{ViT-Small} & Random & 77.4 & \textbf{82.7} & 49.0 \\
  & \baseline{Blockwise} & \baseline{\textbf{77.5}} & \baseline{\textbf{82.7}} & \baseline{\textbf{49.7}} \\
\hline
\multirow{2}{*}{ViT-Base} & Random & \textbf{80.6} & \textbf{85.4} & 52.4 \\
  & \baseline{Blockwise} & \baseline{\textbf{80.6}} & \baseline{85.3} & \baseline{\textbf{52.9}} \\
\hline
\end{tabular}}
\end{center}
\vspace{-0.5cm}
\tabcaption{Ablation on the mask sampling strategy in our \Our. We use the block-wise sampling by default (marked in \colorbox{baselinecolor}{gray}).}
\vspace{-0.7cm}
\label{tab:mask_strategy}
\end{table}

\begin{table*}[t]
\centering
\small
\scalebox{1.0}{
\renewcommand{\arraystretch}{1.05}
\setlength{\tabcolsep}{6.0mm}{
\begin{tabular}{lccccc} %
\toprule
\multirow{2}{*}{Methods} & \multirow{2}{*}{\#Epochs}  &  \multirow{2}{*}{Target} & \multicolumn{2}{c}{IN-1K}  
& \multirow{1}{*}{ADE20K} \\
\cline{4-6}
&  &  & LIN & FT & mIoU \\
\midrule
\midrule
\multicolumn{5}{l}{\emph{Methods using ViT-Tiny}:} \\
MAE-Tiny~\cite{wang2022closer} & 400 & RGB & 23.4 & 76.2 & - \\
CAE~\cite{chen2022context}$^\ddagger$ & 300 & DALL-E & 28.1 & 75.9 & 38.3 \\
Distilled MAE-lite~\cite{wang2022closer} & 400 & RGB & - & 76.5 & - \\
\textbf{\Our} & 300 & CLIP-B & 68.8 & 77.4 & 44.2 \\
\textbf{\Our$+$} & 300 & CLIP-B & \textbf{69.3} & \textbf{77.8} & \textbf{44.7} \\
\midrule
\multicolumn{5}{l}{\emph{Methods using ViT-Small}:} \\
MoCo v3~\cite{chen2021empirical}$^\S$  & 300 & Self-EMA & 73.1 & 81.7$^\dagger$ & - \\
BEiT~\cite{bao2021beit}$^\S$ & 300 & DALL-E & 15.7 & 81.7$^\dagger$ & - \\
SplitMask~\cite{el2021large} & 300  & DALL-E & - & 81.5 & - \\
CAE~\cite{chen2022context} & 300 & DALL-E & 51.8 & 82.0$^\dagger$ & - \\
iBOT~\cite{zhou2021ibot} & 3200 & Self-EMA & \textbf{77.9} & 82.3$^\dagger$ & 45.4 \\
\textbf{\Our} & 300 & CLIP-B & 77.3 & 82.8 & 49.1 \\
\textbf{\Our$+$} & 300 & CLIP-B & 77.5 & \textbf{83.1}$^\dagger$ & \textbf{49.7} \\
\midrule
\multicolumn{5}{l}{\emph{Methods using ViT-Base}:} \\
MoCo v3~\cite{chen2021empirical}  & 300  & Self-EMA & 76.5 & 83.2 & 47.2 \\
DINO~\cite{caron2021emerging}$^\S$ & 400 & Self-EMA & 77.3 & 83.3 & 47.2 \\
iBOT~\cite{zhou2021ibot} & 1600 & Self-EMA & 79.5 & 84.0 & 50.0 \\
BEiT~\cite{bao2021beit}  & 800  & DALL-E & 56.7 & 83.2 & 45.6 \\
SimMIM~\cite{xie2022simmim} & 800 & RGB & 56.7 & 83.8 & - \\
MAE~\cite{he2022masked} & 1600  & RGB & 68.0 & 83.6 & 48.1 \\
CAE~\cite{chen2022context} & 1600  & DALL-E & 70.4 & 83.9 & 50.2 \\
SdAE~\cite{chen2022sdae} & 300  & Self-EMA & 64.9 & 84.1 & 48.6 \\
SIM~\cite{tao2022siamese} & 1600 & Self-EMA & 76.4 & 83.8 & - \\
MaskFeat~\cite{wei2022masked} & 1600  & HOG & - & 84.0 & - \\
SplitMask~\cite{el2021large} & 300  & DALL-E & - & 83.6 & 45.7 \\
PeCo~\cite{dong2021peco} & 800  & VQGAN & - & 84.5 & 48.5 \\
data2vec~\cite{baevski2022data2vec} & 800  & Self-EMA & - & 84.2 & - \\
CMAE~\cite{huang2022contrastive} & 1600 & RGB & - & 84.7 & 50.1 \\
ExtreMA~\cite{wu2022extreme} & 300 & Self-EMA & 73.3 & 83.7 & 47.9 \\
CLIP~\cite{radford2021learning} & -  & Text  & - & 84.9 & 51.1 \\
MaskCLIP \cite{Dong2022MaskCLIPMS} & 1600 & Text & 72.9 & 84.1 & 50.8 \\
MVP~\cite{wei2022mvp} & 300  & CLIP-B & 75.4 & 84.4 & 52.4 \\
BEIT V2~\cite{peng2022beit} & 300 & VQ-CLIP-B & 80.1 & 85.0 & 52.7 \\
\textbf{\Our} & 300 & CLIP-B & 80.5 & 85.2 & \textbf{53.1} \\
\textbf{\Our$+$} & 300 & CLIP-B & \textbf{80.6} & \textbf{85.3} & 52.9 \\
\midrule
\multicolumn{5}{l}{\emph{Methods using ViT-Large}:} \\
MoCo v3~\cite{chen2021empirical}$^\S$  & 300  & Self-EMA & - & 84.1 & 49.1 \\
BEiT~\cite{bao2021beit}$^\S$  & 1600  & DALL-E & - & 85.2 & 53.3 \\
iBOT~\cite{zhou2021ibot} & 1200 & Self-EMA & 81.0 & 84.8 & - \\
MAE~\cite{he2022masked} & 1600  & RGB & 75.8 & 85.9 & 53.6 \\
CAE~\cite{chen2022context} & 1600  & DALL-E & 78.1 & 86.3 & 54.7 \\
data2vec~\cite{baevski2022data2vec} & 1600  & Self-EMA & - & 86.6 & - \\
MVP~\cite{wei2022mvp} & 300  & CLIP-B & - & 86.3 & 54.3 \\
BEIT V2~\cite{peng2022beit} & 300 & VQ-CLIP-B & - & 86.6 & 55.0 \\
\textbf{\Our$+$} & 300 & CLIP-B & \textbf{81.7} & \textbf{86.7} & \textbf{55.9} \\
\bottomrule
\end{tabular}
}
}
\caption{
Pre-training evaluation on the top-1 accuracy (\%) on linear probing (LIN) and fine-tuning (FT) on ImageNet-1K~\cite{deng2009imagenet}, and mIoU (\%) on ADE20K~\cite{zhou2017scene}.
$\dagger$ denotes the fine-tuning epoch is 200 for ViT-Small.
$\ddagger$ means our implementation using the officially released code.
$^\S$ means the results from \cite{chen2022context}.
All other results except for ours are from the original papers.
}
\label{tab:imagenet_seg_sota}
\end{table*}

\myparagraph{Image classification on ImageNet-1K.}
Table~\ref{tab:imagenet_seg_sota} shows the comparisons of different models with two evaluation methods: linear probing and fine-tuning.

With linear probing, \Our shows significant improvements over previous methods with other targets, {\em e.g.}, BEiT~\cite{bao2021beit}, MAE~\cite{he2022masked}, CAE~\cite{chen2022context}, and MaskFeat~\cite{wei2022masked}. These gains are expected as CLIP features contain rich semantics than other targets. Compared with the methods (MVP~\cite{wei2022mvp} and BEIT V2~\cite{peng2022beit}) use CLIP as the target, \Our can also give superior performance (on ViT-Base with 300 epoch pre-training, \Our \vs MVP: 80.6\% \vs 75.4\% and \Our \vs BEIT V2: 80.6\% \vs 80.1\%). When we fine-tune the pre-trained model on ImageNet-1K, \Our achieves the best results among various methods across all scales of ViTs (Table~\ref{tab:imagenet_seg_sota}). Specifically, \Our achieves $\textbf{85.3\%}$ top-$1$ accuracy, surpassing previous methods by large margins. Moreover, with ViT-Large, \Our improves the performance to $\textbf{86.7\%}$ top-$1$ accuracy.

\myparagraph{Semantic segmentation on ADE20K.} 
Semantic segmentation is a challenging task that needs to classify all pixels to various semantic labels given an image. CLIP~\cite{radford2021learning} as the target shows clear advantages in this task. As shown in Table~\ref{tab:imagenet_seg_sota}, \Our significantly improves the results over the methods pre-trained with other targets, {\em e.g.}, by 2.7\% mIoU over CAE~\cite{chen2022context} with ViT-Base. When comparing with CLIP~\cite{radford2021learning}, MVP~\cite{wei2022mvp}, and BEIT V2~\cite{peng2022beit}, \Our outperforms them with the same or less pre-training epochs. The superior performance hold when we move to ViT-Large, with which \Our achieves $\textbf{55.9\%}$ mIoU on ADE20K~\cite{zhou2017scene}.

\myparagraph{Object detection and instance segmentation on COCO.}
We evaluate the pre-trained models on COCO~\cite{lin2014microsoft} with Mask R-CNN~\cite{he2017mask} (Table~\ref{tab:cocodetection_mask}) and Cascade Mask R-CNN~\cite{cai2018cascade,he2017mask} (Table~\ref{tab:cocodetection_cascade}). We report the results of $1\times$ (12 epochs) training schedule.
Compared with other pre-training methods, \Our performs better 
under both two configurations. With Mask R-CNN, \Our gives 4.9/3.0 points higher on ViT-Small and 2.0/0.9 points higher on ViT-Base on AP$^b$/AP$^m$ than the previous best method~\cite{chen2022context}. The superior performance remain when adopting Cascade Mask R-CNN as the fine-tuned model (see Table~\ref{tab:cocodetection_cascade}).

\begin{table}[!t]
    \centering
\small
\renewcommand{\arraystretch}{1.05}
\setlength{\tabcolsep}{4.8mm}{
\begin{tabular}{lccc}
        \toprule
        \multirow{2}{*}{Method} & 
        \multirow{2}{*}{\#Epochs} & 
        \multicolumn{1}{c}{DET}& 
        \multicolumn{1}{c}{INS}\\ 
        \cline{3-4}
        &  & { $\text{AP}^{b}$} & {$\text{AP}^{m}$} \\ 
        \midrule
        \midrule
      \multicolumn{3}{l}{\emph{Methods using ViT-Small}:} \\
       DeiT \cite{touvron2021training}  & 300 & 43.1 & 38.4 \\
       MoCo v3$^*$ \cite{chen2021empirical}  & 300 &  43.3 & 38.8 \\
       BEiT \cite{bao2021beit} & 300 & 35.6 & 32.6 \\
       CAE \cite{chen2022context}  & 300 & 44.1 & 39.2 \\
       \textbf{\Our$+$} & 300  &  \textbf{49.0} & \textbf{42.2} \\
       \midrule
      \multicolumn{3}{l}{\emph{Methods using ViT-Base}:}\\
       DeiT \cite{touvron2021training}  & 300  & 46.9 & 41.5 \\
       MoCo v3$^*$ \cite{chen2021empirical} & 300  & 45.5 & 40.5 \\
       DINO$^*$ \cite{caron2021emerging} & 400 & 46.8 & 41.5 \\
       BEiT \cite{bao2021beit} & 800 & 42.1 & 37.8 \\
       MAE \cite{he2022masked} & 1600 & 48.4 & 42.6 \\
       data2vec~\cite{baevski2022data2vec} & 800 & 41.1 & 37.0 \\
       CAE \cite{chen2022context} & 1600  &  50.0 & 44.0 \\
       \textbf{\Our} & 300  &  51.8 & 44.7 \\
       \textbf{\Our$+$} & 300  &  \textbf{52.0} & \textbf{44.9} \\

        \bottomrule 
    \end{tabular} 
    }
    \caption{Pre-training evaluation on object detection (DET) and instance segmentation (INS) on COCO~\cite{lin2014microsoft}. 
    Mask R-CNN~\cite{he2017mask} is adopted and trained with $1\times$ schedule.
    All results except for \Our are from \cite{chen2022context}.
    \#Epochs refers to the pre-training epochs on ImageNet-$1$K.
    $^*$ denotes multi-crop pre-training augmentation.
    }
    \vspace{-0.5em}
    \label{tab:cocodetection_mask}
\end{table}

\section{Related Work}

Masked image modeling (MIM) aims to learn transferable vision representations. It is inspired by the successful large-scale pre-training for transformers~\cite{vaswani2017attention} with masked language modeling (MLM)~\cite{bert,chen2020generative,gpt3,unilm} in NLP and can serve as a pretext task in self-supervised vision pre-training~\cite{deepcluster, doersch2015unsupervised, cpc, ermolov2020whitening, goyal2021self, li2021prototypical, zbontar2021barlow,moco,mocov2,simclr,byol, Ge2-AE, BootMAE, HiViT, GreenMIM, MimCo, MSN, ConMIM, Li2022mcBEiTMD}. MIM methods~\cite{bao2021beit,he2022masked,xie2022simmim,wei2022masked,baevski2022data2vec, Singh2022RevisitingWS} follow a mask-then-predict pipeline of (i) corrupting an image by masking several image patches based on a pre-defined mask ratio and (ii) learning to predict the missing content under specific supervision. \Our uses the CLIP model~\cite{radford2021learning} as the supervision target and studies on the above two aspects. Next, we discuss related works with respect to these two aspects.

\myparagraph{Supervision target.} There are several ways to represent the missing content when supervising a model. Existing MIM methods explore different supervision targets on their frameworks, including RGB pixels~\cite{he2022masked,gao2022convmae}, HOG descriptors~\cite{wei2022masked}, discrete visual tokens~\cite{bao2021beit,chen2022context,el2021large,peng2022beit,dong2021peco}, and feature representation from momentum models~\cite{tao2022siamese,chen2022sdae,wu2022extreme}. Among these methods, supervision is added to the model's predictions for masked patches. They give no supervision to the model's predictions for unmasked patches (visible patches). Recently, MVP~\cite{wei2022mvp} explored changing the supervision target from other modalities and validated the effectiveness of the additional knowledge. In detail, MVP adopts the vision branch of the CLIP mode\cite{radford2021learning} as the supervision target in MIM. Then MVP gives supervision on all patches of the image, including masked patches and unmasked patches. \Our follows MVP~\cite{wei2022mvp} to use the CLIP model as the supervision target and go one step further to study the supervision position in this paper.

\begin{table}[!t]
    \centering

\small
\renewcommand{\arraystretch}{1.05}
\setlength{\tabcolsep}{4.8mm}{
\begin{tabular}{lccc}
        \toprule
        \multirow{2}{*}{Method} & 
        \multirow{2}{*}{\#Epochs} & 
        \multicolumn{1}{c}{DET}& 
        \multicolumn{1}{c}{INS}\\ 
        \cline{3-4}
        &  & { $\text{AP}^{b}$} & {$\text{AP}^{m}$} \\ 
        \midrule
        \midrule
        
  \multicolumn{3}{l}{\emph{Methods using ViT-Small}:} \\
  iBOT$^*$~\cite{zhou2021ibot} & 3200 & 49.4 & 42.6 \\
  \textbf{\Our$+$} & 300 & \textbf{51.5} & \textbf{43.9} \\
    \midrule
  \multicolumn{3}{l}{\emph{Methods using ViT-Base}:} \\
  MAE \cite{he2022masked} & 1600 & 51.3 & 44.3 \\
  CAE  \cite{chen2022context}  & 1600 & 52.9 & 45.5 \\
  iBOT$^*$~\cite{zhou2021ibot} & 1600 & 51.2 & 44.2 \\
  \textbf{\Our$+$} & 300 & \textbf{53.9} & \textbf{45.9} \\
        \bottomrule 
    \end{tabular} 
    }
    \caption{
    Pre-training evaluation on object detection (DET) and instance segmentation (INS) on COCO~\cite{lin2014microsoft} with Cascade Mask R-CNN~\cite{cai2018cascade}. All models are trained with the $1\times$ schedule.
    All results except for \Our are from \cite{chen2022context},
    and the results of iBOT are from the original paper~\cite{zhou2021ibot}.
    \#Epochs refers to the effective pre-training epochs on ImageNet-$1$K.
    $^*$ denotes the multi-crop pre-training augmentation.
    }
    \vspace{-1.0em}
    \label{tab:cocodetection_cascade}
\end{table}

\myparagraph{Mask ratio.} The mask ratio is a hyper-parameter that needs hand-design in both MLM and MIM. In MLM, BERT~\cite{bert} uses a relatively small mask ratio (15\%) for pre-training. ~\cite{wettig2022should} argues that masking up to 40\% may give higher performance. In MIM, a lot of works use a high mask ratio for pertaining. For example, MAE~\cite{he2022masked} utilizes a mask ratio of
75\%, BEiT~\cite{bao2021beit}, CAE~\cite{chen2022context}, and MVP~\cite{wei2022mvp} empirically set the mask ratio as 40\% and 50\%. In this paper, we study this interesting problem and provide a guideline for choosing the right mask ratio for different scales of ViTs.

Concurrent with our work, \cite{hou2022milan, liu2022exploring, peng2022unified} also explores using CLIP to guide the MIM pre-training. MILAN~\cite{hou2022milan} and dBOT~\cite{liu2022exploring} focus on the impact of target representations, that the CLIP containing multi-modality knowledge can provide more benefits to MIM. MaskDistill~\cite{peng2022unified} works on the design of distillation loss in supervision. Differently, we investigate two aspects orthogonal to these works. We find that adding supervision on visible patches further helps visual learning compared to only supervising the masked patches. Moreover, we explore the relationship between mask ratio and model scales.
These two findings provide useful guidelines for MIM pre-training.

\section{Conclusion and Limitation}
This paper studies two critical ingredients in MIM, \ie, the supervision position and the mask ratio, with CLIP as the supervision target.
With our simple pipeline \Our, we reveal two new insights: i) the feature distillation supervision on visible patches can achieve remarkable performance; ii) the optimal mask ratio is positively correlated to the model size.
Following these two guidelines, our \Our achieves superior performance on all scales of models on various downstream tasks.

\myparagraph{Limitation.} Limited by resources, we do not study on larger models, like ViT-Huge and ViT-Giant. We leave this exploration in the future.

{\small
\bibliographystyle{ieee_fullname}
\bibliography{egbib}
}

\end{document}